\documentclass{article}
\usepackage{spconf,amsmath,graphicx}
\usepackage{todonotes}

\usepackage[utf8]{inputenc} 
\usepackage{cite}
\usepackage{color}
\usepackage{amsfonts}
\usepackage{amsthm}
\usepackage{multirow}
\usepackage{pifont}
\usepackage{url}
\usepackage{stfloats}

\usepackage[T1]{fontenc}  
\usepackage{makecell}
\title{Novel Preprocessing Technique for Data Embedding in Engineering Code Generation Using Large Language Model}
\name{\parbox{16cm}{\centering
        Yu-Chen Lin$^1$\sthanks{This project was completed during the internship at Ansys.}, 
        Akhilesh Kumar$^2$, 
        Norman Chang$^2$,  
        Wenliang Zhang$^2$, \\
        Muhammad Zakir$^2$, 
        Rucha Apte$^2$, 
        Haiyang He$^2$,
        Chao Wang$^2$, 
        Jyh-Shing Roger Jang$^1$
    }
}

\address{
$^1$National Taiwan University, Taiwan\\
$^2$Ansys, Inc., San Jose, California, USA
}

\begin{document}
    %\ninept
    %
    \maketitle
    %
    
    % first version: https://arxiv.org/abs/2311.16267
    \begin{abstract}
        We present four main contributions to enhance the performance of Large Language Models (LLMs) in generating domain-specific code: (i) utilizing LLM-based data splitting and data renovation techniques to improve the semantic representation of embeddings' space; (ii) introducing the Chain of Density for Renovation Credibility (CoDRC), driven by LLMs, and the Adaptive Text Renovation (ATR) algorithm for assessing data renovation reliability; (iii) developing the Implicit Knowledge Expansion and Contemplation (IKEC) Prompt technique; and (iv) effectively refactoring existing scripts to generate new and high-quality scripts with LLMs. By using engineering simulation software RedHawk-SC as a case study, we demonstrate the effectiveness of our data pre-processing method for expanding and categorizing scripts. When combined with IKEC, these techniques enhance the Retrieval-Augmented Generation (RAG) method in retrieving more relevant information, ultimately achieving a 73.33\% "Percentage of Correct Lines" for code generation problems in MapReduce applications.
    \end{abstract}

% Too long?


    \section{Introduction}\label{sec:introduction}
    
    Large Language Models (LLMs) have exhibited remarkable success across diverse applications, showcasing their proficiency in tasks such as domain-specific question answering, code generation, and more \cite{zhao2023survey}. Their adaptability and versatility make them invaluable tools for addressing a wide array of challenges and scenarios. 
    % In recent times, with the rapid rise of Large Language Models (LLMs), both industry and academia have been dedicated to harnessing their powerful capabilities. Specifically, applying LLMs to specialized domains would be a significant achievement, as it would enable LLMs to perform well in data they have never encountered before.
    
    However, applying LLMs for code generation in the RedHawk-SC (RH-SC) domain presents unique challenges. First, generating RH-SC code requires knowledge of circuit design, chip design, and electrical engineering concepts, as well as familiarity with the RH-SC architecture. The documentation for RH-SC is not well-detailed, making it difficult to learn the architecture from the manual, as it assumes readers have a professional background in the field.
    
    The user manual and API documentation are designed for experts in chip design and are not suitable for laypersons, such as LLMs, which may be considered novices in the chip design domain. Furthermore, there is a scarcity of available scripts, making the generation of domain-specific code challenging.
    
    Evaluating the generated code is also difficult due to the nature of the research topic, which focuses on generating code in domains where LLMs have not been trained. This implies that any publicly available information would have already been learned by LLMs, and thus the datasets must be generated by the researchers themselves to avoid any potential biases. To create such datasets, we must utilize the RH-SC tools and system concepts to generate corresponding test code.
    
    Retrieval-Augmented Generation (RAG) \cite{lewis2021retrievalaugmented, gao2024retrievalaugmented} is a commonly used technique that has been proven to reduce hallucinations and enhance the accuracy and reliability of LLMs by fetching facts from external sources. While RAG has achieved notable advancements in specific domains, several challenges still exist.
    % with facts fetched from external sources. 
    
    % When applying LLMs to specific domains, the Retrieval-Augmented Generation (RAG) \cite{lewis2021retrievalaugmented} technique is commonly used, as illustrated in Fig. \ref{fig:standardPipelineInRAG} and \ref{fig:appliedProposedApproachInRAG}. 

    % Data is divided into multiple parts (i.e., chunks) and then converted into embeddings. Simultaneously, the input is also transformed into embeddings, allowing relevant content to be obtained through vector calculations. This content serves as part of the prompt for LLMs, enabling them to respond to domain-specific questions to a certain extent. Consequently, the quality of the retrieved information has a significant impact on RAG's performance, leading to the development of numerous derivative RAG techniques \cite{gao2024retrievalaugmented}.

    RAG's effectiveness lies in the quality of information retrieved, and any shortcomings in this process can directly impact LLM performance and output. One critical concern is the potential for inappropriate data preprocessing, leading to chunks that may not accurately represent their semantic space positions. This discrepancy poses a challenge in effectively locating corresponding content. Even when relevant content is identified, another issue arises: essential information may only constitute a fraction of the retrieved data. Moreover, an excess of information can increase input tokens, potentially giving rise to hallucinations or diluting the importance of crucial details.
    % there exist several challenges.
    % This discrepancy poses a challenge in locating corresponding content effectively.
    
\begin{figure*}[t]
        \centering
        \makebox[\textwidth][c]{\includegraphics[width=0.95\textwidth]{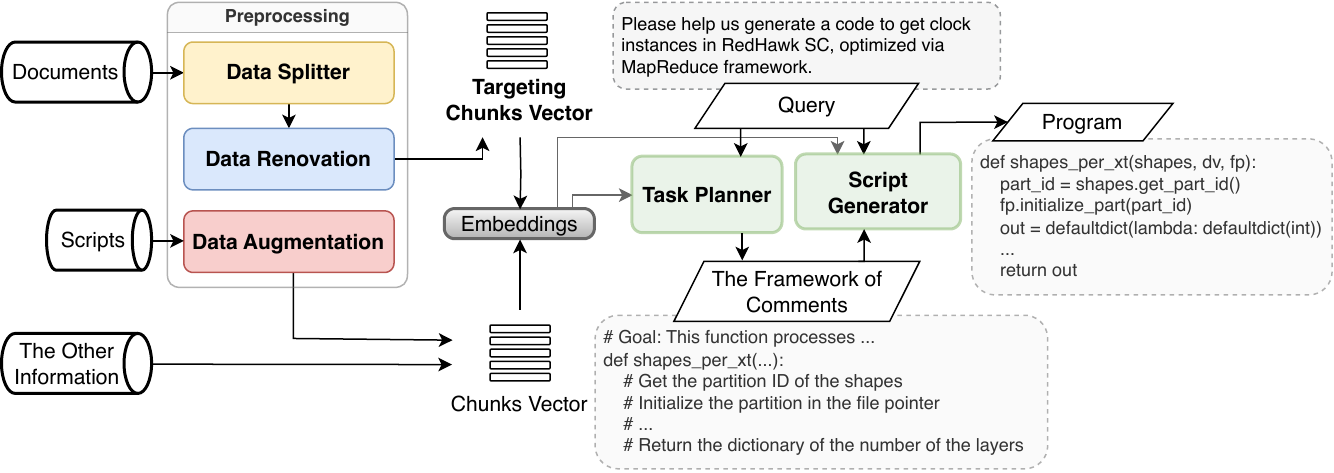}}  
        \caption{Overall flowchart of the code generation process proposed in this study. The "Preprocessing" block includes three innovative data preprocessing techniques introduced in this paper. The Task Planner and Scripts Generator (green rounded rectangles) signify the processing performed by the LLM, combined with the RAG method.}
        \label{fig:overallflowchart}
    \end{figure*}

\begin{figure*}[t]
        \centering
        \makebox[\textwidth][c]{\includegraphics[width=0.95\textwidth]{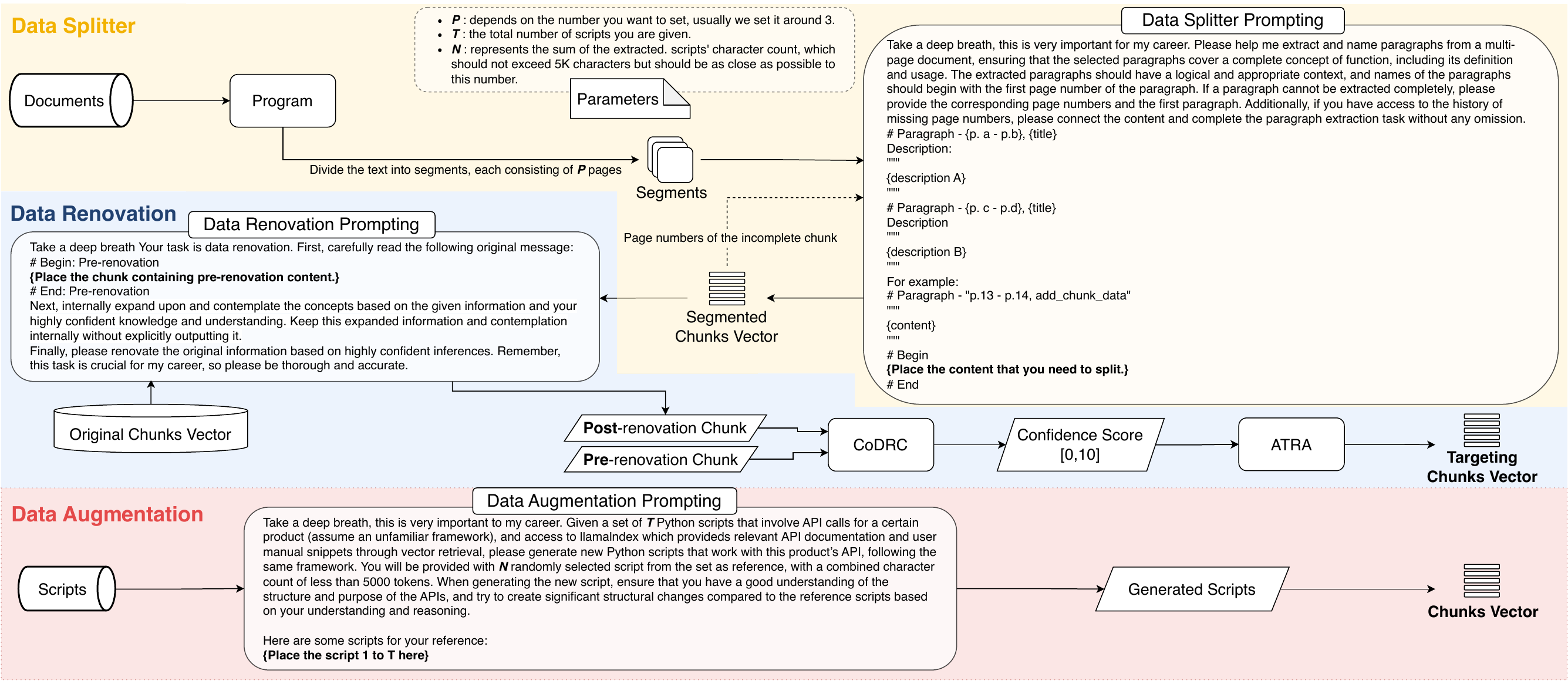}}  
        \caption{Detailed schematic representations of the internal workings of the three methods discussed in this paper (Data Splitter, Data Renovation, Data Augmentation).}
        \label{fig:promptpipeline}
    \end{figure*}

    These challenges underscore the complexities faced during the data preprocessing phase of RAG, emphasizing the need for strategic solutions to enhance its overall efficacy. This paper addresses the critical aspect of data pre-processing by employing a multi-faceted methodology to code generation applications (Fig. \ref{fig:overallflowchart} and Fig. \ref{fig:promptpipeline}). Initially, existing code scripts for RedHawk-SC, an industry-standard tool on dynamic IR analysis, serve as a foundation, enabling LLMs to reconstruct new code scripts and thereby expand the available data.
    
    Following this, LLMs are utilized for optimal segmentation based on semantic paragraphs, and a "renovation" process is applied to each chunk. This addresses the challenge of inaccurate embeddings in technical documents arising from overly concise language.
    
    Simultaneously, a novel method named Chain of Density for Renovation Credibility (CoDRC), derived from the Chain of Density \cite{adams2023sparse}, is introduced. CoDRC is designed to calculate confidence scores after renovation. The process is further refined by incorporating our proposed Adaptive Text Renovation (ATR) algorithm, which approaches the determination of whether to adopt the renovated content from a statistical perspective.
    
    Upon completing the data preprocessing phase and using it as the data source for the RAG method, there is a discernible improvement in RAG's performance. To further enhance the code generation process, the ChatEDA\cite{he2023chateda} code generation workflow is integrated. 
    
    This comprehensive workflow addresses user requirements represented by the Query, which is analogous to the user requirement in ChatEDA, plans the code through the Task Planner, and generates the corresponding code using the Script Generator, as illustrated in Fig. \ref{fig:overallflowchart}. Notably, in our paper's implementation, both the Task Planner and Script Generator are simultaneously combined with the RAG method to acquire relevant knowledge.

    \section{Background}\label{sec:background}
    % Background
    
    In recent years, LLM has rapidly flourished, transforming the entire Natural Language Processing (NLP) field with the rise of ChatGPT. The development of more powerful models, technologies to boost input token limits, more sophisticated prompt techniques, surprisingly powerful distilled models with fewer parameters, and the resurgence of LLM-assisted data processing have all profoundly impacted the world.
    
    \subsection{Models}\label{sec:models}
    Following the release of ChatGPT, closed-source LLMs such as GPT-4-Turbo, Claude, and Gemini have been introduced by various companies, each constantly pushing the boundaries of more powerful and effective LLMs. Open-source LLMs include Vicuna~\cite{peng2023instruction}, Llama2~\cite{touvron2023llama}, and Mistral~\cite{eliseev2023fast}, among others.
    
    Using responses generated by ChatGPT, a dataset of question-and-answer pairs was created for Vicuna. GPT-4~\cite{openai2023gpt4} was then used to evaluate the quality of each question and answer~\cite{zheng2023judging}, marking their performance, and this dataset was utilized as training data for Vicuna. Additionally, a chatbot-arena\footnote{\url{https://huggingface.co/spaces/lmsys/chatbot-arena-leaderboard}} was established, randomly providing two different models to generate answers for the same question, allowing users to choose the better option. This process was treated as a "duel," and after collecting feedback from numerous users, the Elo rating was used to determine the performance priority among multiple models.
    
    % The Vicuna~\cite{peng2023instruction} development team utilized responses generated by ChatGPT to create a dataset of question-and-answer pairs. They then used GPT-4~\cite{openai2023gpt4} to evaluate the quality of each question and answer~\cite{zheng2023judging}, marking their performance and used this dataset as training data for Vicuna. The team further established chatbot-arena~\footnote{\url{https://huggingface.co/spaces/lmsys/chatbot-arena-leaderboard}}, randomly providing two different models to generate answers for the same question, allowing users to choose the better option. This was treated as a "duel," and after collecting feedback from numerous users, the Elo rating was used to determine the performance priority among multiple models.
    % pioneering a novel technique for open-source models.
    
    Mistral~\cite{eliseev2023fast, jiang2024mixtral} employs the "Mixture-of-Experts" (MoE) technique, which consists of multiple experts within a large model, with each expert representing a neural network. This model can distribute token data to different experts internally for individual processing, enabling the training of larger models with lower computational power and further increasing inference speed. In chatbot-arena, Mistral ranked third, only behind GPT-4-Turbo and Bard (Gemini Pro), surpassing ChatGPT as the best-performing open-source model.
    
    Llama2~\cite{touvron2023llama}, an open-source LLM released by Meta, has generated a buzz in both academia and industry due to its powerful functionality and "conditional commercial use." As a result, multiple applications have emerged surrounding this model.

    \subsection{Enhancing Input Tokens} \label{sec:enhancinginputtoken}
    Initially, input tokens posed a significant challenge for LLMs. Since most LLMs are based on the self-attention mechanism~\cite{Vaswani2017attention}, they faced issues with poor efficiency and output quality when dealing with long sequences. However, as the LLM field has flourished, input tokens have gradually become less of a problem. For example, streamingLLM~\cite{xiao2023efficient} proposed an effective method to dynamically spread attention, enabling it to accept an unlimited number of input tokens.
    
    With the gradual resolution of the input token issue, the "dynamic" chunking technique proposed in this paper, which segments text according to paragraphs and semantics, has become a crucial technology for enhancing document search performance.

    \subsection{Retrieval-Augmented Generation (RAG)} \label{sec:rag}
    % Note: Paper review: Retrieval-Augmented Generation for Large Language Models: A Survey
    RAG leverages a multi-step process to enhance information retrieval and subsequent language model responses. The data is initially partitioned into discrete chunks, each of which is then converted into embeddings. Concurrently, the input undergoes a similar transformation into embeddings. This parallel embedding process facilitates the extraction of relevant content through vector calculations. The obtained content serves as a crucial component of the prompt provided to LLMs. By incorporating this tailored information, LLMs are equipped to respond more effectively to domain-specific questions. 
    
    It is noteworthy that the precision and relevance of the retrieved information play a pivotal role in shaping RAG's overall performance. This significance has prompted the development of various derivative RAG techniques aimed at refining and optimizing the process for even better results.
    % Applying the RAG method to LLMs in specialized domains enables them to find relevant text for a corresponding query to serve as reference material, and subsequently answer domain-specific questions. The general approach involves splitting the given domain-specific document into multiple "chunks," converting each into an embedding and storing them in a vector database. On the other hand, user queries are also converted into embeddings, which are used to retrieve related information. This related information is then provided to the LLM as part of the prompt to answer domain-specific questions.
    % The RAG method is commonly applied to LLMs in specialized domains, enabling them to find relevant text for a corresponding query to serve as reference material, and subsequently answer domain-specific questions.
    
    In advanced implementations, the pre-retrieval process can be adjusted, such as enhancing data granularity or optimizing index structure~\cite{gao2024retrievalaugmented}. Embeddings can also be optimized, and specific preprocessing techniques can be applied to documents, along with different segmentation strategies. For example, smaller chunk sizes may work better for some documents, and using certain methods to find the most appropriate chunk size for specific documents is a valuable skill. Additionally, post-retrieval processes can be adjusted, such as prompt compression and re-ranking.

    \subsection{RAG vs Fine-tuning} \label{sec:ragvsfinetuning}
    Using the RAG method, documents can be provided with a certain level of ability to answer domain-specific questions without training a model. In contrast, fine-tuning enables LLMs to "internalize" data as their knowledge when given sufficient information, resulting in outstanding performance in specific domains.
    % The RAG method allows us to provide documents with a certain level of ability to answer domain-specific questions without training a model.
    
    The RAG method does not require abundant computational resources; it only requires domain-specific text-related data and can be directly applied to LLMs, refining prompts without adjusting the LLM. Fine-tuning, on the other hand, requires adequate data and computational power to be performed on open-source LLMs. Currently, closed-source LLMs still outperform their open-source counterparts by a significant margin. However, it is highly likely that using the RAG method and fine-tuning open-source LLMs (e.g., Llama2, CodeGen~\cite{nijkamp2023codegen}) would yield similar performance results, albeit with vastly different costs.

    \subsection{Common Issues with the RAG Method -- Data Splitting} \label{sec:commonissueragmethod}
    
    In the RAG method, we often split data into multiple chunks based on a fixed number of characters, and set an overlap ratio for adjacent chunks~\footnote{\url{https://docs.llamaindex.ai/en/stable/api/llama_index.node_parser.SentenceSplitter.html}} to avoid important information at the chunk's end being cut off directly. 
    
    Although this approach is simple and fast, it cannot effectively segment the data based on semantic, functional, or other meaningful elements. If the actual information needed occupies only a small portion of the corresponding chunk, other content may dominate the chunk. Converting the entire content into an embedding could prevent the semantic space of the chunk from accurately reflecting the position of the required information, making it difficult to find. Even if found, the relevant information might only be a part of the reference material, causing its importance to be diluted and leading to suboptimal outcomes from the self-attention mechanism.
    
    % This approach is simple and fast; however, when examining a single chunk, its composition may consist of elements such as API (a), API (b), and other information.
    
    % If API (a) is the actual content we need, other content may occupy a significant portion of the chunk. Converting the entire content into an embedding may prevent the semantic space of the chunk from accurately reflecting the position of the required information, making it difficult to find. Even if found, the needed information might only be a part of the reference material, causing its importance to be "diluted" and resulting in suboptimal outcomes from the self-attention mechanism.
    
    % \begin{figure*}[t]
    %     \centering
    %     \includegraphics[width=0.90\textwidth]{Images/figure_chunkProblemAndConcept.png}
    %     \caption{Schematic diagram of document segmentation suitable for RAG methods. The left side shows splitting into multiple chunks by C characters with some overlap between them, while the right side illustrates the potential content of a single chunk, such as API (a), API (b) and other information.}
    %     \label{fig:chunkProblemAndConcept}
    % \end{figure*}
    
    To address this issue, existing "Contextual compression~\footnote{\url{https://python.langchain.com/docs/modules/data_connection/retrievers/contextual_compression}}" methods have been proposed, such as removing "unhelpful" words before text segmentation and retrieving more related text, then extracting only the helpful content based on query relevance. However, these methods may either damage the original meaning of the text or disrupt word structures, leading to "illusions" and failing to address the problem fundamentally.

    \subsection{Prompt Techniques and Mechanisms} \label{sec:promptechandmechanism}
    
    To achieve better performance with prompt mechanisms, previous approaches such as Chain-of-Thought (CoT)~\cite{wei2022chain} have been proposed, which involves providing examples and step-by-step processes to enhance performance. Zero-Shot Chain of Thought, on the other hand, prompts the model to output a sequence of thoughts step by step to boost performance. 
    
    Later, to solve more complex problems, ReAct~\cite{yao2022react} was developed, breaking down the original question into simpler sub-questions, answering them one by one, and ultimately combining all the information to answer the original question, thereby improving performance on complex problems. Other approaches include "Successive Prompting" ~\cite{dua2022successive} and "Take a Step Back,"~\cite{zheng2023step} which trace the question back to its underlying "theorem" or "concept" before answering the original question to enhance performance.
    
    On the other hand, techniques within the prompt itself have also been explored, such as using "emotional blackmail"~\cite{li2023large} or the "IKEC" method mentioned in this paper to boost performance.

    \subsection{Distilling} \label{sec:distilling}
    
    The distilling approach involves utilizing a large LLM to initially generate rationales for a given answer in the training set, and then employing those rationales to train a smaller model~\cite{hsieh2023distilling}. By reducing the data size, the number of model parameters can be decreased, enabling smaller model parameters to maintain existing performance with minimal impact while also reducing the overall number of parameters. In a similar vein, Microsoft's Orca2~\cite{mitra2023orca} also follows this direction, striving to achieve the goal of creating more compact models.
    
    % The distilling approach requires a large LLM to first generate rationales for a given answer in the training set and then use that to train the smaller model ~\cite{hsieh2023distilling}. Reducing the data size allows for a decrease in model parameters, and smaller model parameters can maintain the existing performance with minimal impact while reducing model parameters. Microsoft's Orca2~\cite{mitra2023orca} is no exception in pursuing the goal of creating more compact models.

    % \subsection{Self-Planning Code Generation with LLMs} \label{sec:selfplanningllm}
    % % Ref: Paper
    % In this approach~\cite{jiang2023selfplanning}, a problem statement is first provided, followed by the generation of multiple subtasks, and then the corresponding code for each task is generated sequentially. By utilizing a "progressive" generation method, the performance of algorithm-related problems is significantly improved.
    
    \subsection{Related Work in Code Generation} \label{sec:relatedwork}
    This section presents an overview of recent advancements and techniques in code generation using LLMs and their applications in specific domains.
    
    In the domain of code generation, noteworthy related works include self-planning code generation with LLMs~\cite{jiang2023selfplanning}, the Chain of Code (CoC)~\cite{li2023chain} approach, and the AgentCoder~\cite{huang2024agentcoder} framework. Self-planning code generation employs a progressive generation strategy by dividing tasks into multiple subtasks, significantly improving the performance of algorithm-related problems. The CoC approach encourages the formatting of semantic subtasks in a program as flexible pseudocode, which leads to significant improvements in LLM performance for logic and arithmetic tasks.
    
    Additionally, the Active Retrieval Augmented Generation~\cite{jiang2023active}, with its proposed technique called Forward-Looking Active REtrieval (FLARE), emphasizes actively deciding when and what to retrieve across the course of the generation, demonstrating effectiveness in various long-form, knowledge-intensive generation tasks. The AgentCoder framework provides a multi-agent system that iteratively improves code generation by developing and testing code based on feedback, surpassing the limitations of single-agent models and traditional methodologies.
    
    ChatEDA~\cite{he2023chateda} has achieved significant success in code generation within the EDA domain, using fine-tuning on Llama2 and obtaining excellent results with the Self-Planning Code Generation with LLMs~\cite{jiang2023selfplanning} method. TestPilot~\cite{schäfer2023empirical}, on the other hand, focuses on adjusting the LLM application process and Prompt Engineering without fine-tuning, also achieving commendable results in code generation for the Mocha Framework. VeriGen~\cite{thakur2023verigen} employs online Verilog-related code and textbooks for fine-tuning the CodeGen-16B model while testing its performance with three different levels of prompt detail.
    
    The last three approaches concentrate on code generation application research within specific domains. ChatEDA has a unique method for data preparation, utilizing limited code to allow LLMs to reassemble multiple times to create an Instruction Pool. The data is then checked semi-manually before being used for fine-tuning. VeriGen, on the other hand, uses online resources for fine-tuning, as Verilog has slightly more code resources. TestPilot, with many more resources available for Mocha compared to the other two, can achieve excellent results by simply adjusting the process and prompt, considering that LLM already has partial knowledge.
    
    Building on the aforementioned related literature, we considered whether good results could be achieved without fine-tuning by simply improving information retrieval. Therefore, we referred to ChatEDA's code generation process and data augmentation and proposed our data preprocessing method to enhance the relevance of information retrieval.

    \section{Methodology}\label{sec:methodology}
    
    Fig. \ref{fig:overallflowchart} presents the overall flowchart of the code generation process in this paper, demonstrating how our proposed preprocessing method integrates with the RAG technique. The framework of comments is generated by the Task Planner, and this result, along with the Query, is used to iteratively produce corresponding code. Fig. \ref{fig:promptpipeline} delves into the internal mechanism of our approach, including document processing, Prompts, and more. The bold text should be replaced with context-specific content, and the highlighted sentences are the core sentences of IKEC. We will subsequently explain each component in greater detail.

    \subsection{Data Augmentation} \label{subsec:dataaugmentation}
    
    % \begin{figure*}[t]
    %     \centering
    %     \includegraphics[width=0.90\textwidth]{Images/figure_dataAugmentationPrompt.png}
    %     \caption{Schematic diagram of Data Augmentation Prompt, where the actual prompt content is shown. Red characters represent the main keywords, bold text indicates key sentences in the prompt, and yellow text refers to key sentences from related literature that help improve LLM performance.}
    %     \label{fig:dataAugmentationPrompt}
    % \end{figure*}
    
    We randomly select several scripts (typically two to three) from the original $T$ scripts, ensuring the total number of characters does not exceed $C_s$ (set to 5000 in this paper). Then, as shown in the corresponding block in Fig. \ref{fig:promptpipeline} (with a pink background), we place the content in the Prompt sequentially and combine it with the RAG method. At this point, the data source for the RAG method is all the documents that have not yet undergone our proposed preprocessing technique, which is first split into several chunks by characters and then converted into embeddings.
    
    In the Prompt, we use the key sentence "\textit{Take a deep breath, this is very important to my career}"  from previous research, which has been proven to effectively improve performance \cite{yang2023large, li2023large}, and this sentence will appear repeatedly in subsequent Prompts. We then inform the model of its task and background. Notably, we indicate that the model must have a good understanding of the structure and purpose of APIs, and generate the scripts based on this understanding and reasoning. Emphasizing this point improves the rationality of the generated scripts.
    
    Another key aspect is hinting at "significant structure changes." If we can reconstruct scripts with vastly different structures based on the original scripts, it would be a relatively simple and helpful approach. Conversely, if we ask the LLM to generate high-quality "new" scripts based on the original scripts and provide related text, it is much more challenging. This is a major difference in our method. Generating more scripts as reference data sources will greatly benefit the performance of RAG.

    \subsection{Implicit Knowledge Expansion and Contemplation (IKEC)} \label{subsec:IKEC}
    
    In previous research, the Zero-shot Chain of Thought method enabled LLMs to output their thought process step by step, thus enhancing their performance. Scratchpads \cite{nye2021show}, on the other hand, added detailed thought processes to the provided examples, helping LLMs understand the examples and perform better. However, both methods increase the number of input and output tokens, as well as the cost and inference time.
    
    The main concept of our uniquely developed Prompt technique, IKEC, is to encourage the model to internally expand and supplement content it is highly confident about, based on its knowledge, carefully ponder, and only then generate a response without outputting the process. This approach can improve performance without increasing the number of input and output tokens, particularly in effectively avoiding fatal logic errors in code. A demonstration of this technique can be seen in the example provided in Fig. \ref{fig:ikecresult}.
    
    We specifically apply this technique to Data Renovation and Data Augmentation, mainly because both require obtaining expert knowledge and producing accurate results under good mastery. Especially for the renovated data and scripts, it is crucial to generate content as reliable as possible. % Utilizing the IKEC mechanism allows for the creation of reliable content with the highest confidence.

    % IKEC is a novel prompt technique introduced in this paper. Previously, the Zero-shot Chain of Thought method encouraged LLM to output its thought process, thereby improving its performance. Scratchpads \cite{nye2021show}, on the other hand, provided examples of the thought process. Both of these approaches increase input tokens and output tokens.
    % 
    % The main concept of IKEC is to encourage the LLM to think and expand upon highly confident content based on its existing knowledge, contemplate it internally, and then solve the problem without outputting these processes. This approach can improve performance without increasing input tokens and output tokens, saving costs and reducing inference time.
    % 
    % In this paper, IKEC is specifically applied to "Data Renovation." The reason for this is that renovation requires obtaining expert knowledge of content and providing as reliable content as possible after renovation. IKEC stimulates the LLM to expand and supplement the most confident content internally based on the information obtained and its own knowledge, which can help Data Renovation generate reliable content.

    \subsection{Data Splitter} \label{subsec:datasplitter}
    
    Upon obtaining the corresponding Segment, we use the Prompt shown in Fig. \ref{fig:promptpipeline} (with a yellow background) and place the entire content of the Segment in the bold text section, allowing it to be split into segmented chunks vector semantically according to the customized rules.
    
    The Prompt briefly explains the task, informs the model of the rules to follow during the splitting, provides format and actual example references, and finally presents the Segment that needs to be split.
    
    This splitting method ensures that each chunk has a more focused and single-topic or concept "dynamic" length, which better reflects the true semantic spatial position without disrupting the original content, making it easier for RAG to find the most relevant and helpful chunk content.
    
    Splitting is essentially a "binary classification problem," evaluating whether to split at each position. Therefore, compared to generating text, splitting is a relatively simple task and falls within the familiar domain of LLMs. It is reasonable to assume that LLMs can perform well in this task, as it is easier for them compared to generating answers. When splitting is done effectively, this can help enhance the performance of RAG.
    
    % Splitting is essentially a "binary classification problem," evaluating whether to split at each position. Therefore, compared to generating text, splitting is a relatively simple task and falls within the familiar domain of LLMs. We firmly believe that LLMs can perform well in this task, and this simple operation can effectively boost the performance of RAG.
    
    It is worth noting that since we provide text data on a "page number" basis, the first paragraph of the first page and the last paragraph of the last page are likely to be incomplete. Therefore, we let the LLM output the "incomplete paragraph" with the corresponding page number as a historical record for the next iteration, allowing it to combine the incomplete paragraph with the first paragraph of the next content before splitting.
    
    % \begin{figure*}[t]
    %     \centering
    %     \includegraphics[width=0.90\textwidth]{Images/figure_dataSplitterPrompt.png}
    %     \caption{Schematic diagram of Data Splitter Prompt}
    %     \label{fig:dataSplitterPrompt}
    % \end{figure*}
    
    % Data Splitter is a "binary classification problem", which means that for each position, the LLM needs to predict whether the position requires splitting or not. Therefore, compared to generating text, this kind of problem is relatively simpler for the LLM.
    % 
    % The Data Splitter obtains the corresponding text and splits it into multiple paragraphs according to custom rules, as shown in the prompts in Fig. \ref{fig:dataSplitterPrompt}. Initially, a brief description of the task is given, followed by the customized rules, a format, and actual examples. Finally, the text to be split is provided, completing the prompt.
    % 
    % It is worth noting that since we provide text data page by page, the first paragraph on the first page and the last paragraph on the last page are likely to be incomplete. Therefore, we ask LLM to record the "incomplete paragraphs" and use them as historical records for the next iteration, allowing it to combine the incomplete paragraphs with the first paragraph of the next content.
    % 
    % This method of segmentation allows for "dynamic" length chunks, ensuring that each chunk has a single main theme or concept, as opposed to "fixed character length" chunks previously. In this way, the chunks can better reflect their true semantic space positions without damaging the original content.

    \subsection{Data Renovation} \label{subsec:datarenovation}
    
    After obtaining each Chunk individually, we use the Prompt shown in Fig. \ref{fig:promptpipeline} (with a blue background). We first inform the model of its task and provide the Chunk content that needs to be renovated. Then, we apply the IKEC Prompt technique (fluorescently highlighted text) and ultimately prompt the model to infer based on the highly mastered content and thoroughly consider generating accurate content.
    
    Data Renovation mitigates the issue of technical documents generally being written concisely, allowing LLMs to renovate each Chunk one by one, providing reliable and more detailed content. The confidence level of the content before and after renovation is assessed using CoDRC, and the final decision to adopt the renovated content is made in conjunction with the ATR Algorithm, ensuring the quality of the renovated content.
    
    If the documents are not renovated and the content is excessively concise, even if the LLM can obtain the corresponding text, it is prone to produce poor content with limited understanding, significantly affecting its performance. Moreover, more detailed information will help improve the accuracy of Embedding, making it easier to find the corresponding Chunk. This is the reason why we use Data Renovation.

    % \begin{figure*}[t]
    %     \centering
    %     \includegraphics[width=0.90\textwidth]{Images/figure_dataRenovationPrompt.png}
    %     \caption{Schematic diagram of Data Renovation Prompt, where red text represents the main key sentences of IKEC, and yellow text refers to key sentences from related literature that help improve the effect of the prompt.}
    %     \label{fig:dataRenovationPrompt}
    % \end{figure*}
    % 
    % After Data Splitter has divided the text into individual chunks, Data Renovation obtains each chunk in sequence and renovates it using IKEC, as shown in the prompt in Fig. \ref{fig:dataRenovationPrompt}. Initially, the task is introduced, and the content of the chunk to be renovated is provided. Then, IKEC is used, and finally, the LLM is instructed to output well-thought-out and accurate content based on highly credible inferences.
    % 
    % Data Renovation is necessary because, in most technical documents, the content is written concisely and precisely for a professional audience. This can affect LLM's performance to some extent, even when it has access to the corresponding text, and it may not understand the content at all. Furthermore, because concise content is shorter, its semantic space may be less accurate.
    % 
    % Therefore, we hope to have LLM renovate each chunk one by one, providing the "most reliable" renovated content. Subsequently, the CoDRC and ATRA algorithms are used to decide whether to renovate, ensuring the quality of the content.

    \subsection{Chain of Density for Renovation Credibility (CoDRC)} \label{subsec:CoDRC}
    Chain of Density \cite{adams2023sparse} is often used in translation or summarization tasks. It identifies "Informative Entities" and "Missing Entities" in two steps and, combined with Guidelines, enables the model to achieve better quality based on previous results, even approaching human performance.
    
    % Chain of Density (CoD) \cite{adams2023sparse} is often used in translation or summarization tasks. It identifies "Informative Entities" and "Missing Entities" in two steps and combines them with guidelines to achieve better quality based on previous results, even approaching human performance.
    
    We modify the steps of CoD and apply it to the assessment of Renovation Credibility. Firstly, "Renovation" usually results in more detailed and meticulous Post-renovation content, implying more words. Applying the original CoD to the context of summarization tasks, we can treat "Post-renovation" as the "pre-summarization" content and "Pre-renovation" as the "post-summarization" content. We then capture the corresponding "Missing Entities" using the following method:
    
    % In this paper, we modified the steps of CoD and applied them to the evaluation of Renovation Credibility. Firstly, "Renovation" usually results in more detailed and meticulous post-renovation content, implying a higher word count. 
    
    % In the context of the original CoD application for summarization tasks, we can consider the "Post-renovation" content as the "pre-summarization" content, and the "Pre-renovation" content as the "post-summarization" content. By doing so, we can capture the corresponding "Missing Entities" with the following method:

    \begin{itemize}
        \item Step 1: Given the renovation history record, request the Post-renovation content, Pre-renovation content (treated as content after summarization), and any other information obtained.
        \item Step 2: Ask the LLM to compare Pre-renovation and Post-renovation and list all missing messages (Informative entities) that were not mentioned in the Pre-renovation content but were mentioned in the Post-renovation content. Categorize them as follows:
        \begin{enumerate}
            \item Can be inferred based on previously obtained information.
            \item Can be inferred based on relevant technical background and knowledge.
            \item Neither of the above, i.e., information that cannot be inferred through relevant knowledge, background, or expertise.
        \end{enumerate}
        \item Step 3: Based on the above content, ask the LLM to give a confidence score (0-10 points) based on this information. The higher the score, the more information can be reasonably inferred. Then, save both the Pre-renovation and Post-renovation information, and record the confidence score within the content of the "Post-renovation" information, so that subsequent processes can evaluate and decide whether to adopt the content before or after renovation.
        % Then, record both the Pre-renovation and Post-renovation information in a txt file and write the score in the Post-renovation file name.
    \end{itemize}
    
    % \begin{itemize}
    %     \item Step 1: Given the renovation history record, request the Post-renovation content, Pre-renovation content (treated as content after summarization), and any other information obtained.
    %     \item Step 2: Ask to compare Pre-renovation and Post-renovation and list all missing messages (Informative entities) that were not mentioned in the Pre-renovation content but were mentioned in the Post-renovation content. Categorize them as follows:
    %         \begin{enumerate}
    %             \item Can be inferred based on previously obtained information.  
    % 		\item Can be inferred based on relevant technical background and knowledge.  
    % 		\item Neither of the above, i.e., information that cannot be inferred through relevant knowledge, background, or expertise.
    %         \end{enumerate}
    %     \item Step 3: Based on the above content, ask the LLM to give a score (0-10 points) based on this information. The higher the score, the more information can be reasonably inferred. Then, record both the Pre-renovation and Post-renovation information in a .txt file and write the score in the Post-renovation file name. 
    % \end{itemize}
    
    We apply part of the CoD method to identify "Missing Entities" in the content before and after renovation, which means the added content after renovation, and categorize it. Finally, based on the categorization results, pre and post-renovation information, and additional content provided, the LLM is asked to give a confidence score $\text{Conf}_i$ ranging from 0 to 10.
    
    % We applied parts of the CoD method to identify the "Missing Entities" in both pre- and post-renovation content, which refers to the newly added content after renovation. Furthermore, we classified these entities. Ultimately, based on the classification results, pre- and post-renovation information, and the provided additional content, the LLM assigns a confidence score $\text{Conf}_i$ ranging from 0 to 10.
    
    % We applied parts of the CoD method to identify the "Missing Entities" in the pre- and post-renovation content, which implies the newly added content after renovation. Furthermore, we classified these entities. Ultimately, based on the classification results, pre- and post-renovation information, and the provided additional content, the LLM assigns a confidence score $\text{Conf}_i$ ranging from 0 to 10.
    
    Using this method to select added content and categorize how the information is obtained makes the confidence score assessment easier and more reliable and serves as an essential indicator of whether to adopt the renovated content.

    \subsection{Adaptive Text Renovation Algorithm (ATRA)} \label{subsec:ATRA}
    
    Following the previously mentioned CoDRC method, we have a "confidence score" $\text{Conf}_i$ ranging from zero to ten, with values up to the first decimal place, for all post-renovation content. We define a function $\text{Count}(\text{parameter})$ that returns the total number of characters in $\text{parameter}$. Next, we define the "Text Growth Ratio" $\text{Grow}_i$ as follows:
    % Following the previously mentioned CoDRC method, we have a "confidence score" $\text{Conf}_i$ between zero to ten for all post-renovation content. We define a function $\text{Count}(\text{parameter})$ that returns the total number of characters in $\text{parameter}$. Next, we define the "Text Growth Ratio" $\text{Grow}_i$ as follows:
    
    \begin{equation}
        \text{Grow}_i = \frac{\text{Count}(\text{Post.}) - \text{Count}(\text{Pre.}) }{\text{Count}(\text{Pre.})}
        \label{eq:grow}
    \end{equation}

    Now we can obtain the corresponding $\text{Conf}i$ and $\text{Grow}i$ for each post-renovation content. Therefore, for the "confidence score," we can calculate its mean $\overline{\text{Conf}}$ and standard deviation $\sigma_\text{Conf}$. Similarly, for the "Text Growth Ratio," we can calculate its mean $\overline{\text{Grow}}$ and standard deviation $\sigma_{\text{Grow}}$.
    
    Next, we calculate the following formula for each post-renovation content:
    
    \begin{equation}
        \text{Cdiff}_i = \frac{\text{Conf}_i - \overline{\text{Conf}}}{\sigma_{\text{Conf}}}
        \label{eq:cdiff}
    \end{equation}
    
    The above formula calculates how many standard deviations the post-renovation information is above or below the mean value. Similarly, the "Text Growth Ratio" is calculated in the same way:
    
    \begin{equation}
       \text{Gdiff}_i = \frac{\text{Grow}_i - \overline{\text{Grow}}}{\sigma_{\text{Grow}}} 
       \label{eq:gdiff}
    \end{equation}
    
    Given $\text{Cdiff}_i$ and $\text{Gdiff}_i$, we can evaluate the renovation based on Equation (\ref{eq:diffcompare})
    % Next, we only need to compare the two values:
    
    \begin{equation}
      \text{Cdiff}_i - \text{Gdiff}_i \geq \text{Constant}\footnote{Constant is a user-defined arbitrary rational number."}
      \label{eq:diffcompare}
    \end{equation}
    
    The $\text{Constant}$ in the previous formula can be adjusted according to the actual situation. With this calculation method, the entire document serves as a reference for adjustment. The more text added after renovation, the higher the required confidence level. The $\text{Constant}$ serves as the threshold; when the threshold is higher, a higher confidence score is needed to renovate the data, whereas a lower threshold allows for data renovation with a lower confidence score.
    % The $\text{Constant}$ is used to adjust the "threshold"; when it is positive, the threshold is higher, and when it is negative, the threshold is lower.
    
    We make use of the balance between the confidence score $\text{Conf}_i$ and the text growth ratio $\text{Grow}_i$ to retain as much reliable post-renovation content as possible.

    %%%%%%%%%%%%%%%%%%%%
    % Overall for Approach
    \par
    
    In the above sub-sections, we have explained in detail the methods and concepts of each key technique. For documents, we have obtained the Targeting Chunks Vector, which are Chunks with more focused and detailed topic content. For Scripts, we have reconstructed new Scripts based on the original ones to expand the data, while the remaining information is left untouched. Finally, we can convert all mixed Chunks Vectors into Embeddings one by one, making them suitable for the RAG method.

    \subsection{Code Generator} \label{subsec:codegenerator}
    
\begin{figure*}[t]
        \centering
        \includegraphics[width=0.95\textwidth]{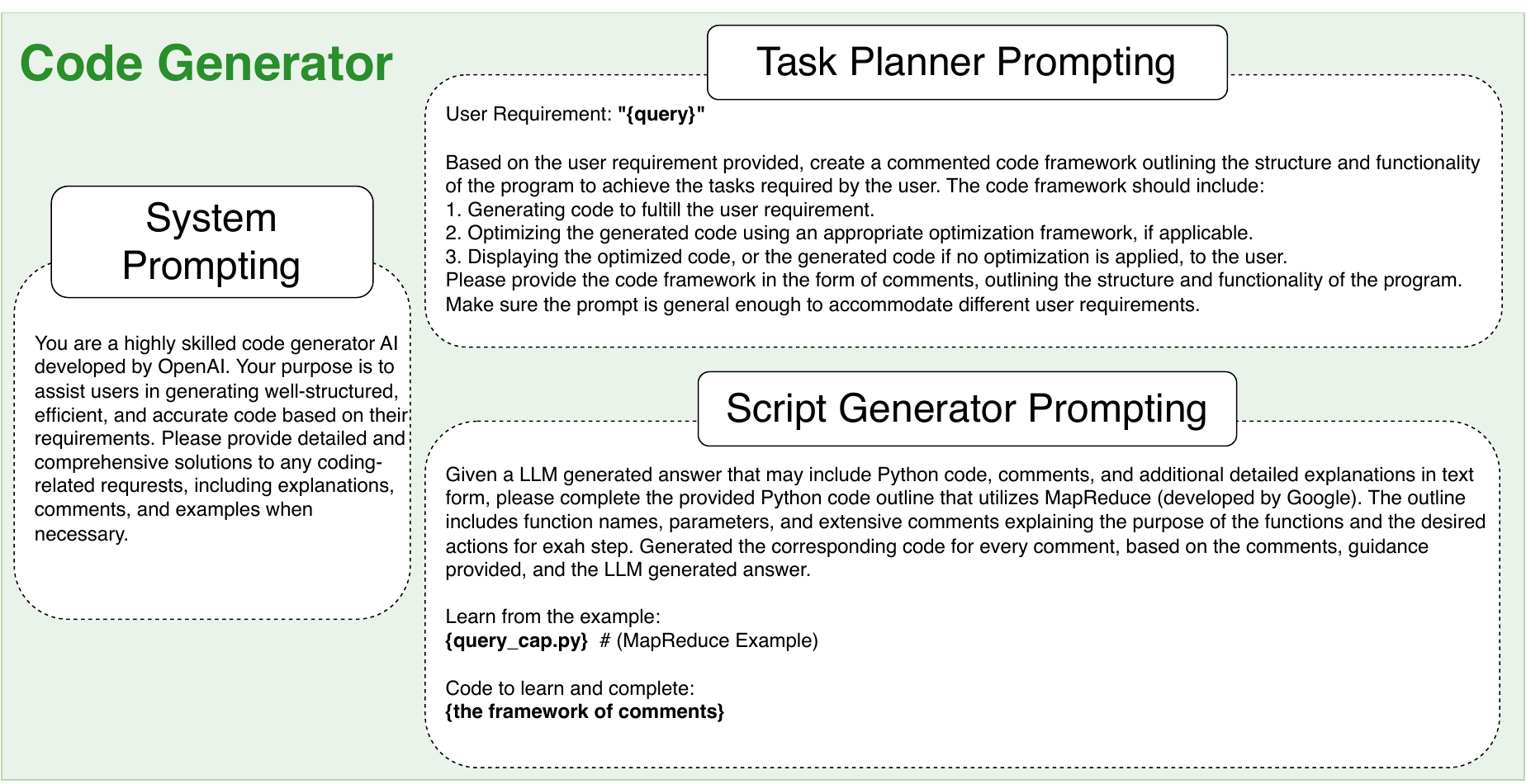}
        \caption{Code Generator Prompt - The Prompt used in the code generation process with the same System Prompt. The bold text should be replaced with the corresponding content, such as replacing "query" with the user's required sentence.}
        \label{fig:codegeneratorprompt}
    \end{figure*}

    After obtaining the Targeting Chunk Vector, we use it as the data source for the RAG method. Then, following the steps of generating code in ChatEDA \cite{he2023chateda} (Fig. \ref{fig:overallflowchart}), (i) Query: Provide user requirements and use the System Prompt in the following steps, (ii) Task Planner: Use the prompt for LLM with the RAG method to plan the outline of the code and then generate the framework of comments, (iii) Script Generator: Use the prompt for LLM with the RAG method to generate the code one by one according to the corresponding comments.  All prompts in these steps refer to Fig. \ref{fig:codegeneratorprompt}. This process ultimately produces domain-specific code based on the Query.
    
    % After completing the data preprocessing, we refer to the ChatEDA \cite{he2023chateda} paper's methodology (Fig. \ref{fig:codeGeneratorPipeline}) and perform the following steps: (1) provide user requirements, (2) use an LLM to generate an outline of the code design in the form of annotations, (3) pass the results from step (2) to generate code corresponding to each annotation, and (4) produce the corresponding code.
    
    % Initially, provide the user requirements (query) and use the System Prompt (Fig. \ref{fig:systemprompt}) and Task Planner Prompt (Fig. \ref{fig:taskPlannerprompt}) to obtain the code outline generated in the form of comments. Then, use the Script Generator Prompt (Fig. \ref{fig:scriptgeneratorprompt}) to provide a MapReduce example and the results obtained from the Task Planner to generate the corresponding code. Finally, obtain the program tailored to the user's requirements.
    
    % \begin{figure}[t]
    %     \centering
    %     \includegraphics[width=0.45\textwidth]{Images/figure_system_prompt.png}
    %     \caption{Code Generation - System Prompt. This is the unified System Prompt used by the Task Planner and Script Generator.}
    %     \label{fig:systemprompt}
    % \end{figure}
    
    % \begin{figure}[t]
    %     \centering
    %     \includegraphics[width=0.45\textwidth]{Images/figure_taskPlanner_prompt.png}
    %     \caption{Task Planner Prompt - Replace the foundation with the corresponding user requirements, as illustrated by the query in Fig. \ref{fig:overallflowchart}).}
    %     \label{fig:taskPlannerprompt}
    % \end{figure}
    % 
    % \begin{figure}[t]
    %     \centering
    %     \includegraphics[width=0.45\textwidth]{Images/figure_scriptGenerator_prompt.png}
    %     \caption{Script Generator Prompt - MapReduce Example. Place any script using MapReduce here, and place the framework of comments generated by the Task Planner in the Code (Only Comment) section.}
    %     \label{fig:scriptgeneratorprompt}
    % \end{figure}

    \section{Experiment}\label{sec:experiment}
    
    We take the generation of code for the RedHawk-SC engineering simulation tool and the acceleration of simulation speed using MapReduce~\cite{dean2008mapreduce} as examples of domain-specific code generation. Our experiments are divided into two different types:
    
    % In our experiments, we use the company's RedHawk-SC as an example of domain-specific code generation. We conduct three different types of experiments:
    
    \begin{itemize}
        \item Type A: Individual testing of various components, such as Data Augmentation, Data Splitter, Data Renovation, and Implicit Knowledge Expansion and Contemplation (IKEC), each demonstrated using single example outcomes.
        \item Type B: A comparison of the performance of different code generation methods.
        % \item Type 1: Test Data Augmentation, Data Splitter, Data Renovation, and IKEC individually, presenting the results with a "single example."
        % \item Type 2: Test whether the preprocessing methods described in this paper can effectively improve "accuracy."
        % \item Type 3: Test the effects of the methods described in this paper on Task Planner and Script Generator.
    \end{itemize}
    
    In this paper, the Type B experiments have only preliminary results available, with complete findings yet to be obtained. These will be elaborated on in the following sub-sections.
    
    % It is worth noting that the Type 2 and 3 experiments mentioned in this paper are currently only experimental designs and do not yet have complete results.

    \subsection{Type A - Individual Tests}\label{subsec:individualtests}
    
    In the individual tests, we individually assess the effectiveness of each key technique presented in this paper, showcasing them through actual experimental results. For a detailed description of each component, please refer to the Methodology section.
    
    % In individual tests, we examine the effects of each key technique presented in this paper and demonstrate them using actual experimental results.
    
    \subsubsection{Data Augmentation} \label{subsubsec:dataaugmentation}
    
    \begin{figure*}[t]
        \centering
        \includegraphics[width=0.95\textwidth]{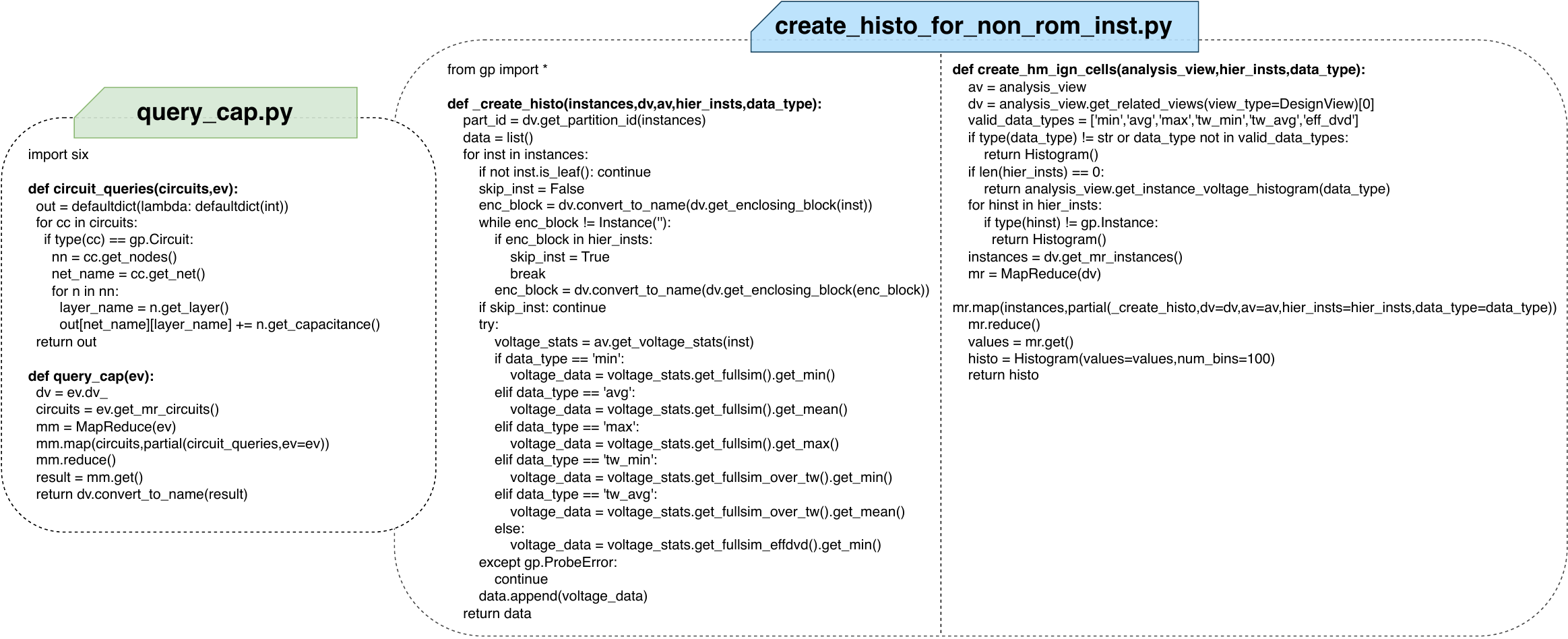}
        \caption{Data Augmentation Example - The source code of the two Scripts used.}
        \label{fig:sourcecodedataaugmentation}
    \end{figure*}
    
    \begin{figure*}[t]
        \centering
        \includegraphics[width=0.95\textwidth]{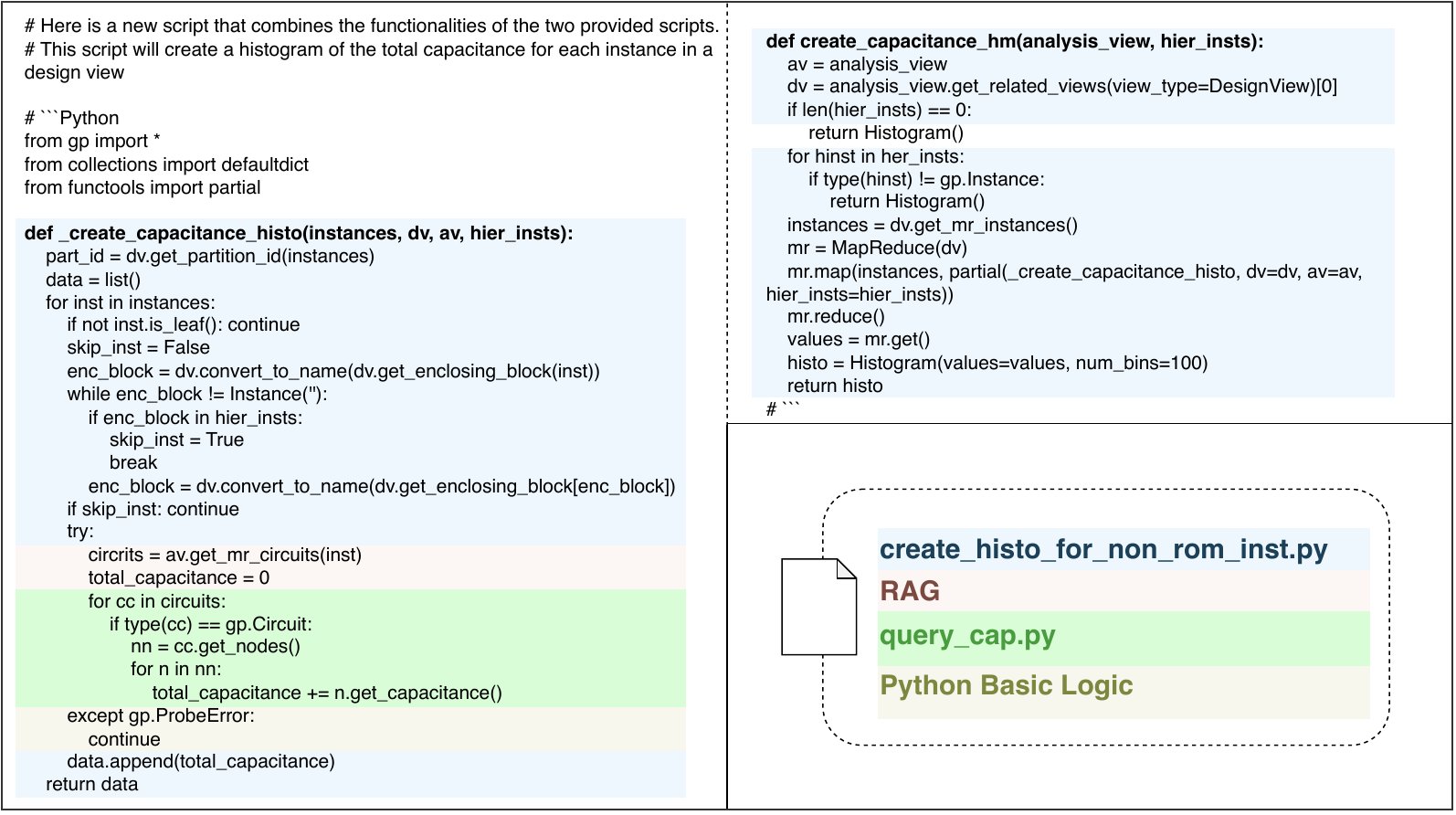}
        \caption{Example (one of the experimental results demonstration) - Data Augmentation, a new script generated using the above two source codes and applying the LLM method. (The light blue and green backgrounds represent parts of the code with a structure similar to the corresponding colored source code in Fig. \ref{fig:sourcecodedataaugmentation}, pink indicates code generated from data obtained by referencing the RAG method, and yellow denotes code generated using basic Python logic.)}
        \label{fig:dataAugmentationexperiment}
    \end{figure*}

    We use the prompt shown in Fig. \ref{fig:promptpipeline} (with a pink background) and arbitrarily select two scripts (Fig. \ref{fig:sourcecodedataaugmentation}) for testing. These are generated using GPT-4 (temperature = 0, other parameters: default). As illustrated in Fig. \ref{fig:dataAugmentationexperiment}, we can see that it reassembles the original scripts while combining the information obtained from RAG and fundamental Python logic to reconstruct a syntactically smooth script that aligns with its intended goal.
    
    % We use the Prompt shown in Fig. \ref{fig:dataAugmentationPrompt} and randomly select two scripts (Fig. \ref{fig:querycap}: query\_cap.py, Fig. \ref{fig:createhistofornonrominst}: create\_histo\_for\_non\_rom\_inst.py) for testing. The actual results are shown in Fig. \ref{fig:dataAugmentationexperiment}, with the meanings of the lines in the image explained in its caption. We can see that it can reassemble the original script and, combined with the information from RAG and Python's basic syntax, reconstruct a script with smooth syntax and in line with its intended purpose.
    
    Following this method, we repeat the process on the existing 23 scripts, generating 20 new scripts, which are then manually evaluated by in-house RedHawk-SC experts. Experimental results reveal that all syntax is correct, and the code is generated following the LLM's self-defined objectives. However, optimizations in code performance are still needed.
    
    % Following this method, we internally use the original 23 scripts and generate 20 new scripts, which are then manually evaluated by our Application Engineering team. The syntax and purpose of all generated scripts are found to be accurate.

    \subsubsection{Data Splitter} \label{subsubsec:datasplitter}
    
    \begin{figure*}[t]
        \centering
        \includegraphics[width=0.95\textwidth]{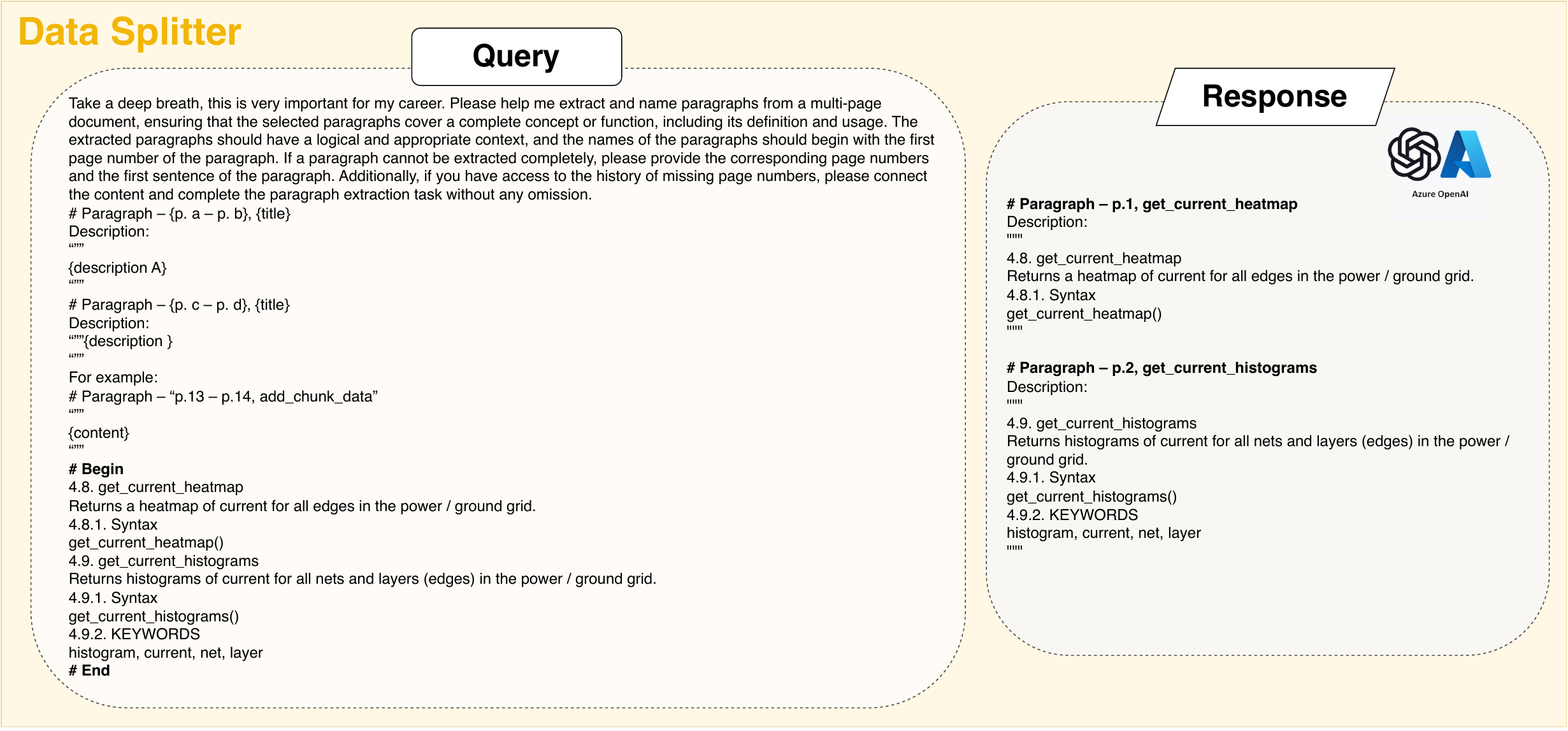}
        \caption{Example - Actual experimental results of the Data Splitter, using a portion of API documentation}
        \label{fig:dataSplitterExperiment}
    \end{figure*}
    
    We used a portion of the API documentation as an example (Fig. \ref{fig:dataSplitterExperiment}). As seen on the right side of the figure, the Response accurately segments the input into different chunks based on the requested format, such as "get\_current\_heatmap" and "get\_current\_histograms," resulting in proper separation according to function names.
    
    % In the Data Splitter experiment, we used a portion of the API documentation as an example (Fig. \ref{fig:dataSplitterExperiment}). As seen on the right side of the figure, it correctly segments the text into different chunks according to the function format.
    
    The test environment shown in the figure is Azure OpenAI Studio (parameters: GPT-4, temperature = 0.0, other parameters: default), which is also used in subsequent tests.
    
    % The environment used for the image is Azure OpenAI Studio, with the parameter model set to gpt-4 and temperature set to 0.0. All other settings are kept as default. The same applies to the following Data Renovation section.

    \subsubsection{Data Renovation} \label{subsubsec:datarenovation}
    
    \begin{figure*}[t]
        \centering
        \includegraphics[width=0.95\textwidth]{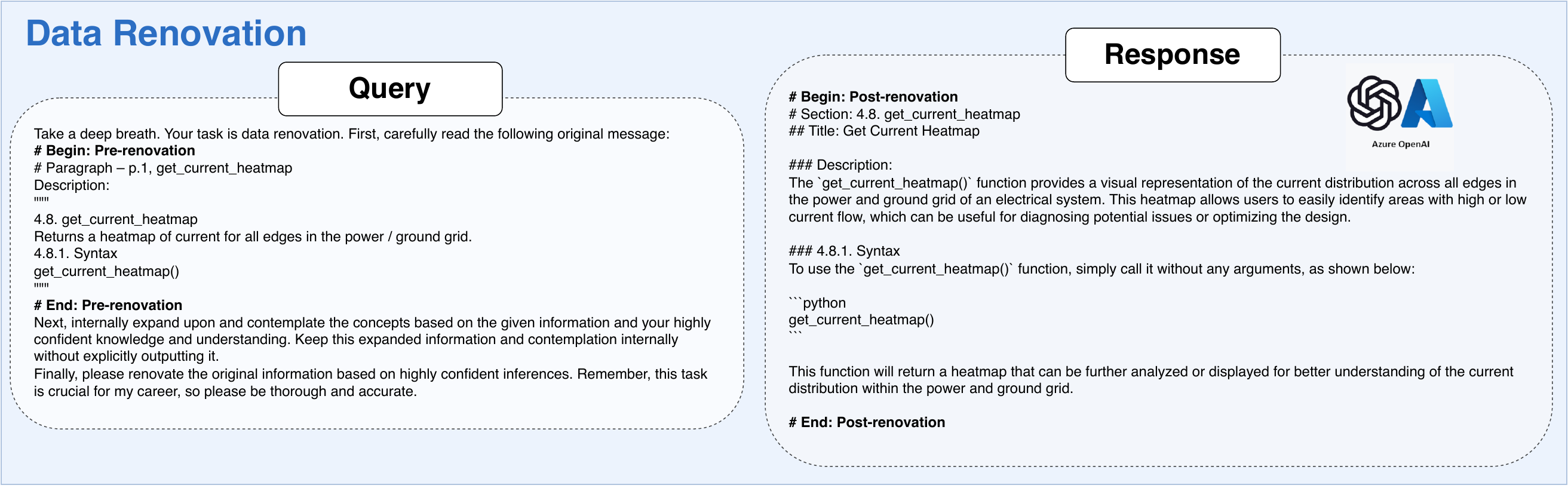}
        \caption{Example - Results of the Data Renovation experiment - using Chunks obtained from the Data Splitter.}
        \label{fig:dataRenovationExperiment}
    \end{figure*}
    
    % \begin{figure}[t]
    %     \centering
    %     \includegraphics[width=0.45\textwidth]{Images/figure_dataRenovationExperimentResponse.png}
    %     \caption{Actual experiment response for Data Renovation, shown as a screenshot of the Azure OpenAI Studio interface. The only adjusted parameter is "temperature," set to 0.0, while the rest remain at their default values.}
    %     \label{fig:dataRenovationExperimentResponse}
    % \end{figure}
    We then use the Chunk: "get\_current\_heatmap" generated by the Data Splitter as an example (Fig. \ref{fig:dataRenovationExperiment}). As shown in the result on the right, the original description of the function is quite brief. Even without using RAG, the LLM can still infer accurate content based on its existing knowledge and the Chunk's content while supplementing parameter usage and return values. After renovation, additional reliable information is provided, enhancing the existing content. When combined with the RAG method and subsequent CoDRC and ATR Algorithm evaluations, this approach further ensures the reliability of the renovated content.

    % In this experiment, we use the results generated by the previously mentioned Data Splitter. We employ the query shown in Fig. \ref{fig:dataRenovationExperimentQuery}, with different colored blocks corresponding to the different colored blocks in Fig. \ref{fig:dataRenovationPrompt}. The obtained results are shown in Fig. \ref{fig:dataRenovationExperimentResponse}. We used the first chunk from the Data Splitter example, "Add Chunk Data." From the actual renovated results, we can see that it does not modify the existing text. Instead, it supplements the parameter definitions and usage methods, striving to provide more reliable content.
    
    In the individual tests, we individually assess the effectiveness of each key technique presented in this paper, showcasing them through actual experimental results. 
    % For a detailed description of each component, please refer to the Methodology section.

    \subsubsection{Implicit Knowledge Expansion and Contemplation (IKEC)} \label{subsubsec:ikec}
    
    \begin{figure}[t]
        \centering
        \includegraphics[width=0.45\textwidth]{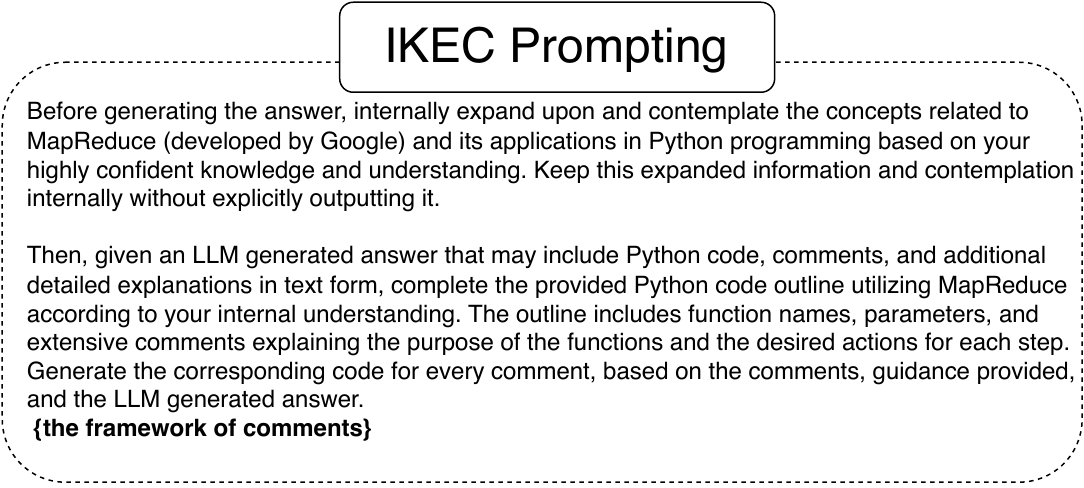}
        % \caption{Example of IKEC - Prompt, this is an example of generating code using IKEC technique in Task Planner. Bold text represents the keywords used during IKEC, the yellow text indicates the most important key sentences in IKEC method, and the last section "Only Comment" refers to the content of the code outline written in a comment form.}
        \caption{IKEC Example Prompt - An example of generating code using IKEC in the Script Generator. The first paragraph applies the core concept of IKEC, while the bold text at the end contains the framework of comments generated by the Task Planner.}
        \label{fig:promptikec}
    \end{figure}
    
    \begin{figure*}[t]
        \centering
        \includegraphics[width=0.95\textwidth]{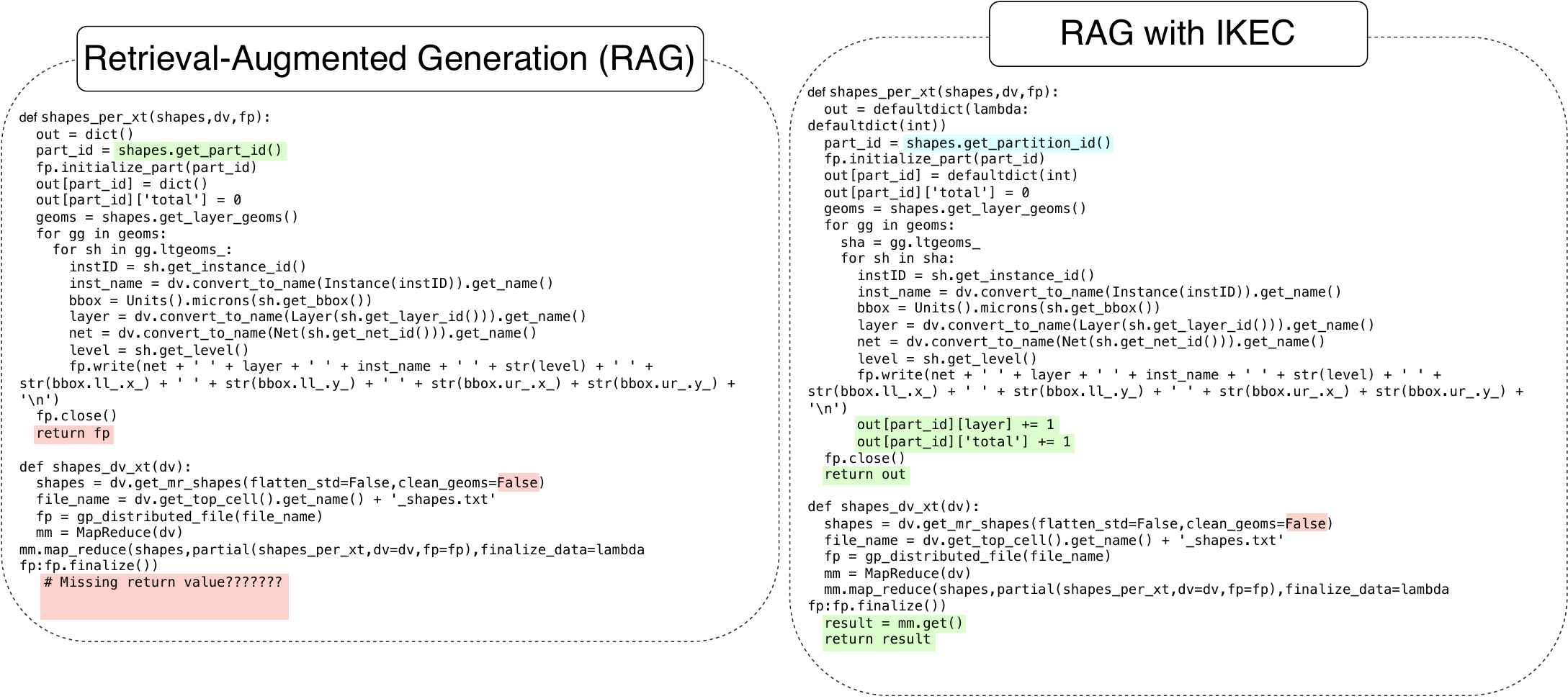}
        \caption{Example of IKEC - Experimental Results, these are the results of using Fig. \ref{fig:promptikec}. On the left side is the general RAG method, and on the right side is RAG + IKEC. A green background indicates correct, a red background signifies obvious errors or omissions, and a light blue background denotes function name errors.}
        \label{fig:ikecresult}
    \end{figure*}

    Although IKEC is primarily applied to Data Renovation and Data Augmentation, we have designed separate experiments to verify its practical effects (Prompt as shown in Fig. \ref{fig:promptikec}). When given the same framework of comments, the results with and without using IKEC are displayed in Fig. \ref{fig:ikecresult}.
    
    % Although IKEC has been directly applied to Data Renovation, we have used the same technique to test "code generation." We use the Prompt shown in Fig. \ref{fig:promptikec}, and the obtained results are displayed in Fig. \ref{fig:ikecresult}.
    
    The first function needs to calculate the total number of layers, while the second function requires MapReduce to accelerate the layer calculation of the prior function. On the left side, when using the RAG method directly, the first function omits the code for calculating the number of layers and does not return a dictionary result; the second function fails to return the MapReduce result. 
    
    In contrast, the right side shows the use of IKEC in conjunction with the RAG method, resulting in significant improvements. The number of layers is accurately calculated, returning the corresponding dictionary as well as the MapReduce result. The only error is a misnamed function call, which can be improved by applying the preprocessing techniques described in this paper.
    
    % The first generated function needs to calculate the number of layers, while the second function needs to apply MapReduce to speed up the calculation of the first function. As seen on the left side, the results obtained directly using the RAG method do not calculate the number of layers in the first function, nor do they return the dictionary result. The second function does not return the MapReduce result either. These issues are significantly improved after using IKEC (right side), which correctly calculates the number of layers and returns the corresponding dictionary, as well as the MapReduce result.

    \subsection{Type B - Performance Test for Code Generation}\label{subsec:performancetestforcodegeneration}
    % \subsection{Type 2 - Accuracy Experiment}\label{subsec:accuracyexperiment}
    
    Some preliminary experiments have been carried out to test the performance of code generation with the proposed workflow. We currently have preliminary experimental results and a comprehensive experimental design. However, it is important to note that the full experiment has not yet been completed.
    % The performance test for code generation directly assesses the performance of the generated code.
    
    \subsubsection{Preliminary Experiment}
    Initially, we selected five scripts as subjects, treating each as a standard answer, and manually generated corresponding queries (i.e., User Requirements) for them. We then used only RAG in combination with the IKEC Prompt technique to generate code following the ChatEDA process, first creating the framework of comments and then generating the code.
    
    The evaluation method involves manually marking errors in the generated code, such as incorrect function names and syntax errors. The "Percentage of Correct Lines" is calculated by dividing the number of lines without errors by the total number of lines.
    
    We use the example of IKEC in Fig. \ref{fig:ikecresult}: The total number of correct lines in the code is twenty-nine. On the left side, using the RAG method directly, we can see three erroneous lines, among which "Missing return value" actually represents two missing lines (correctly shown on the right). Therefore, the percentage of correct lines on the left side is calculated as $\frac{29 - 4}{29} \approx 86.21 \%$. Similarly, for the version using the IKEC technique on the right side, we can see one red-marked error and one blue function name error, totaling two lines, resulting in a calculation of $\frac{29 - 2}{29} \approx 93.10 \%$.
    
    Using this calculation method, the final results are averaged across five generated code samples, yielding a 73.33\% "Percentage of Correct Lines".

    \subsubsection{Complete Experimental Design}
    Next, we plan to use twenty scripts as a foundation, treating each as a standard answer and manually generating corresponding queries for them. Additionally, we will manually assign difficulty scores (0-10) and categorize them as "easy" (0-4), "medium" (5-7), and "hard" (8-10). The difficulty definitions are as follows:
    \begin{itemize}
        \item Easy: only requires correct retrieval of the corresponding text.
        \item Medium: requires correct retrieval of the corresponding text, some knowledge of RedHawk-SC, and basic logic.
        \item Hard: requires correct retrieval of corresponding text from multiple different chapters and familiarity with RedHawk-SC concepts.
    \end{itemize}
    
    Subsequently, internal RedHawk-SC experts will design unit tests for each script to detect functionality and syntax errors. The evaluation criteria are as follows:
    \begin{itemize}
        \item Executability: If the code has no syntax errors and can be executed, it is considered correct; otherwise, it is considered incorrect.
        \item Functionality: Based on the unit tests designed by the experts, different scripts may have varying numbers of unit tests. We will calculate the pass rate of unit tests for each script and take the average in the end.
    \end{itemize}
    
    Ultimately, we will design the following experiments, with a particular emphasis on ablation study comparisons:
    \begin{itemize}
        \item Baseline - RAG: Directly generate code based on the query using the RAG method.
        \item Baseline - ReAct: Directly generate code based on the query using the ReAct method.
        % \item Baseline - RAG + Semantic Chunker: This method, proposed by Greg Kamradt on llamaIndex, splits chunks based on semantics. After preprocessing using this method, generate code directly based on the query using the RAG method.
        \item RAG + Data Splitter: First, use the Data Splitter method described in this paper to split the data into chunks, then generate code using the RAG method.
        \item Baseline - RAG + ChatEDA: Use the RAG method to first generate the framework of comments based on the query, then produce code.
        \item Baseline - ReAct + ChatEDA: Use ReAct to first generate the framework of comments based on the query, then produce code.
        % \item Baseline - RAG + Semantic Chunker + ChatEDA: This method, proposed by Greg Kamradt on llamaIndex, splits chunks based on semantics. After preprocessing using this method, generate the framework of comments based on the query using the RAG method, and then produce code.
        \item RAG + Data Splitter + ChatEDA: First, use the Data Splitter method described in this paper to split the data into chunks, then generate the framework of comments using the RAG method, and finally produce code.
        \item RAG + Data Splitter + Renovation: Use the data preprocessing method in this paper, excluding CoDRC and ATR Algorithm. Finally, generate the code directly using the RAG method.
        \item RAG + Data Splitter + Renovation (without CoDRC and ATRA) + ChatEDA: Use the data preprocessing method in this paper, excluding CoDRC and ATR Algorithm. Then, generate the framework of comments using the RAG method, and finally produce code.
        \item RAG + Data Splitter + Renovation: Use the data preprocessing method in this paper, and then generate the code directly using the RAG method.
        % Use the data preprocessing method in this paper, excluding CoDRC and ATR Algorithm. Finally, generate the code directly using the RAG method.
        % Perform data preprocessing as described in this paper, then generate code using the RAG method.
        % \item RAG + Data Splitter + Renovation (CoDRC + ATRA): Perform data preprocessing as described in this paper, then generate code using the RAG method.
        \item \textbf{Our proposed (RAG + Data Splitter + Renovation + ChatEDA)}: Perform data preprocessing as described in this paper, then generate the framework of comments using the RAG method, and finally produce code.
        % \item RAG + Data Splitter + Renovation + ChatEDA: Use the data preprocessing method in this paper, then generate the framework of comments using the RAG method, and finally produce code.
        % lack renovation, renovation + CoDRC & ATR Algorithm. let's ignore it first
        % IKEC experiment?
        % lack ReAct if time permit...
    \end{itemize}
    
    In total, there are ten experimental groups. Additionally, for all methods using RAG, we will adjust the parameters "chunk\_size" (the maximum total number of words per chunk) and "similarity\_top\_k" (the number of reference chunks taken in each operation). Other settings, such as temperature, are fixed at 0, and we will use GPT-4/3.5 for the experiments, with all other parameters set to their default values. If the experiments go smoothly, we will expand the number of topics to 40 and increase the testing of different LLMs.

    \section{Conclusion}\label{sec:conclusion}

    Domain-specific code generation presents a challenging yet promising arena. Achieving this, as outlined in our paper, involves providing just the right text without the need for additional fine-tuning. Such a method promises rapid deployment across numerous fields. Notably, advancements in "algorithm design" code generation could significantly revolutionize the LLM domain.
    
    Our proposed method makes significant contributions, particularly in the area of data preprocessing. This paper is the first to suggest the use of LLM for "data segmentation" and "data renovation" to enhance embedding accuracy. Furthermore, we propose an effective data augmentation technique that generates additional high-quality Scripts from the original ones, providing valuable reference sources and contributing to the improvement of the RAG method's performance. Our proposed IKEC method also stimulates deep thinking within the model, leading to better performance. In our preliminary experiments, we utilized the ChatEDA workflow combined with IKEC-generated code and achieved a 73.33\% "Percentage of Correct Lines" for code generation problems in MapReduce applications. However, experimental results for our overall method presented in this paper are yet to be obtained.

    \section{Acknowledgment}\label{sec:acknowledgement}

    This paper extends special thanks to Ansys Inc. for providing resources and to Ansys Fellow Norman Chang for their assistance and guidance on this project. We are particularly grateful to Jibin John for providing over twenty original scripts and to Wenliang Zhang for helping to review the generated code and annotate the original scripts. Akhilesh Kumar, Rucha Apte, Haiyang He and Chao Wang have recently been engaged in similar projects and have contributed invaluable insights during our meetings. Additionally, Muhammad Zakir and the aforementioned colleagues have also assisted me in running code with the company's products.
    
    The method proposed in this paper won the championship among 18 teams in the Generative AI category at the Tech New Stars competition, organized by the Department of Industrial Technology (DOIT), Ministry of Economic Affairs (MOEA), R.O.C., Taiwan. We are very grateful to the organizers for recognizing the potential of the method described in this paper and affirming it with the championship.
    
    Finally, we would also like to thank Jyh-Shing Roger Jang and all the students of the Natural Language Processing (NLP) group at the National Taiwan University's MIRlab (Multimedia Information Retrieval Laboratory) for their frequent and collaborative discussions.
    
    There are so many people to thank; we are deeply grateful to all the personnel involved in this project!

    \section{Statement}\label{sec:statement}

    This paper aims to present preliminary results and ideas, contributing to the academic community's development in the field of Large Language Models (LLMs). However, certain aspects, such as evaluation, have not yet been fully realized. Most significant improvements have been observed in small datasets (at least five questions), and we plan to continue refining and updating the findings in this paper.
    
    \bibliographystyle{IEEEbib}
    \bibliography{arXiv2023}

    % \import{./Sections/}{8-appendix}
\end{document}